\documentclass[journal]{IEEEtran}

\usepackage{graphicx}
\usepackage{amsmath,amssymb} 
\usepackage{color}
 \usepackage{bbm, dsfont}
\usepackage{multirow}
 
\usepackage{longtable}
\usepackage[T1]{fontenc}
\usepackage{epsfig}
\usepackage{authblk}
\usepackage[symbol*]{footmisc}
\usepackage{floatrow}
\usepackage{caption}
\usepackage{hyperref}

\usepackage{mathtools}
\usepackage{resizegather}
\usepackage{dsfont}
\usepackage{caption}
\usepackage{multirow}

\usepackage{amsthm}
\theoremstyle{plain}

\usepackage[section]{placeins}

\newfloatcommand{capbtabbox}{table}[][\FBwidth]
\usepackage[table]{xcolor}
\definecolor{city_color_0}{rgb}{0.0,0.0,0.0}
\definecolor{city_color_1}{rgb}{0.5020,0.2510,0.5020}
\definecolor{city_color_2}{rgb}{0.9569,0.1373,0.9098}
\definecolor{city_color_3}{rgb}{0.2745,0.2745,0.2745}
\definecolor{city_color_4}{rgb}{0.4000,0.4000,0.6118}
\definecolor{city_color_5}{rgb}{0.7451,0.6000,0.6000}
\definecolor{city_color_6}{rgb}{0.6000,0.6000,0.6000}
\definecolor{city_color_7}{rgb}{0.9804,0.6667,0.1176}
\definecolor{city_color_8}{rgb}{0.8627,0.8627,0.0000}
\definecolor{city_color_9}{rgb}{0.4196,0.5569,0.1373}
\definecolor{city_color_10}{rgb}{0.5961,0.9843,0.5961}
\definecolor{city_color_11}{rgb}{0.2745,0.5098,0.7059}
\definecolor{city_color_12}{rgb}{0.8627,0.0784,0.2353}
\definecolor{city_color_13}{rgb}{1.0000,0.0000,0.0000}
\definecolor{city_color_14}{rgb}{0.0000,0.0000,0.5569}
\definecolor{city_color_15}{rgb}{0.0000,0.0000,0.2745}
\definecolor{city_color_16}{rgb}{0.0000,0.2353,0.3922}
\definecolor{city_color_17}{rgb}{0.0000,0.3137,0.3922}
\definecolor{city_color_18}{rgb}{0.0000,0.0000,0.9020}
\definecolor{city_color_19}{rgb}{0.4667,0.0431,0.1255}

\usepackage{epsfig}
\usepackage{amsmath,amssymb} 
\usepackage{multirow}

\begin{document}
 
\title{Reinforced Wasserstein Training for Severity-Aware Semantic Segmentation in Autonomous Driving}

\author{Xiaofeng~Liu,
        Yimeng Zhang,
        Xiongchang Liu,
        Song Bai,
        Site Li,
        and~Jane~You
\thanks{Xiaofeng Liu is with the Harvard University, Cambridge, MA, 02138 USA (Corresponding author: xliu11@bidmc.harvard.edu).}
\thanks{Yimeng Zhang is with the Columbia University and Harvard University, Cambridge, MA, 02138 USA.}
\thanks{Xiongchang Liu is with the China University of Mining and Technology and Harvard University, Cambridge, MA, 02138 USA.}
\thanks{S. Bai is with the Department
of Statistics, the University of California Berkeley, Berkeley, CA, 94720 USA.}
\thanks{Site Li is with the Carnegie Mellon University, Pittsburgh,
PA, 15232 USA.}
\thanks{J. You is with the Dept.of Computing, The Hong Kong Polytechnic University, Hong Kong.}
\thanks{Accepted to IEEE T-ITS}}

\maketitle

\begin{abstract}
Semantic segmentation is important for many real-world systems, e.g., autonomous vehicles, which predict the class of each pixel. Recently, deep networks achieved significant progress w.r.t. the mean Intersection-over Union (mIoU) with the cross-entropy loss. However, the cross entropy loss can essentially ignore the difference of severity for an autonomous car with different wrong prediction mistakes. For example, predicting the car to the road is much more servery than recognize it as the bus. Targeting for this difficulty, we develop a Wasserstein training framework to explore the inter-class correlation by defining its ground metric as misclassification severity. The ground metric of Wasserstein distance can be pre-defined following the experience on a specific task. From the optimization perspective, we further propose to set the ground metric as an increasing function of the pre-defined ground metric. Furthermore, an adaptively learning scheme of the ground matrix is proposed to utilize the high-fidelity CARLA simulator. Specifically, we follow a reinforcement alternative learning scheme. The experiments on both CamVid and Cityscapes datasets evidenced the effectiveness of our Wasserstein loss. The SegNet, ENet, FCN and Deeplab networks can be adapted following a plug in manner. We achieve significant improves on the predefined important classes, and much longer continuous play time in our simulator.
\end{abstract}


\IEEEpeerreviewmaketitle

\section{Introduction}

\IEEEPARstart{S}emantic segmentation (SS) has been an important computer vision task, which aiming to densely predict the discrete class labels of the pixel of image \cite{xiang2019comparative,yang2019pass}. For an autonomous driving, robotics, augmented reality and automatic surgery system, it is an important way to precisely understand the scene. Recently, many work have been done in this area \cite{zou2019confidence,alvarez2010road},
and leading to considerable progress on major open benchmark datasets \cite{cordts2016cityscapes} with the advances of deep learning technology. In the deep learning era \cite{liu2020auto3d,liu2019dependency,He_2020_CVPR_Workshops,He_2020_CVPR_Workshops,liu2019permutation,liu2019feature}, segmentation is essentially making the pixel-wise classification based on cross-entropy (CE) loss.

Unfortunately, the aforementioned models can encounter challenges in many practical applications such as autonomous driving, where one has different severity w.r.t. different misclassification cases. For example, an accident of Tesla is caused by a wrong recognition of a white truck as sky, arousing intense discussion of autonomous vehicle safety\footnote{\url{https://www.nytimes.com/2017/01/19/business/tesla-model-s-autopilot-fatal-crash.html}}. However, the result may have been different had just recognized the truck as car/bus. Similarly, Uber's car misclassified a person and finally resulted in a pedestrian being killed\footnote{\url{https://nypost.com/2019/11/07/}}.

\begin{figure}[t]
\centering
\includegraphics[width=8cm]{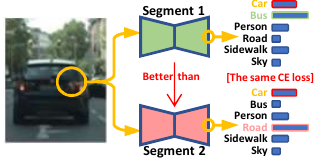}\\
\caption{The limitation of CE loss for real-world autonomous driving system. The true class of these pixels are car ${i^*}$. We show two softmax output of the segmenters which have the same probability at ${i^*}$ position. They will be assigned with the same cross-entropy loss, while the first distribution can be more preferable than the second one. These two results can lead to different severity consequences.}\label{fig:1} 
\end{figure}

As illustrated in Fig. \ref{fig:1}, compared with the bottom segmentation prediction (Car$\rightarrow$Road), the top one is more preferable (Car$\rightarrow$Bus), while the CE loss does not discriminate these two softmax probability histograms. We note that with one-hot ground-truth label, the CE loss is only related to the prediction probability of the true class $p_{i^*}$, where ${i^*}$ is the index of the true class. More formally, $\mathcal{L}_{CE}=-{\rm log}p_{i^*}$. 

Actually, there are severity correlations of each label classes $e,g.,$ severity(Car$\rightarrow$Bus)$>$severity(Car$\rightarrow$road) and severity(Person$\rightarrow$Road)$>$severity(Sky$\rightarrow$Road). When using the cross-entropy objective, the classes are independent to each other \cite{liu2018ordinal}, and the inter-class relationships are not been considered.

Our claim is also closely related to the importance-aware classification/segmentation \cite{chen2018importance,xiang2019importance}. These methods were proposed to define some class groups based on the pre-defined importance of each class. For example, the car, truck, bus are in the most important group, road and sidewalks are in the less important group, and the sky is in the least important group. Then, a larger weight will be multiplied to the more important group to calculate the loss. Therefore, misclassifying a car as $any$ other classes will receive larger punishment than misclassifying the sky as $any$ other classes. This is a nice property, but not sufficient for safe driving as it cannot discriminate the severity of different prediction in misclassification cases. $e.g.,$ Fig. \ref{fig:1}.

Targeting the aforementioned difficulties, we choose the Wasserstein distance as the alternative optimization objective. The first order Wasserstein distance can be regard as the optimal transport cost of moving the mass in one distribution to match the target distribution \cite{ruschendorf1985wasserstein}. In this paper, we propose to calculate the Wasserstein distance between the softmax prediction of segmentor and its target label. We note that both of them are the normalized histograms. By setting the ground matrix as the misclassification severity, we are able to measure the prediction that sensitive to the pair-wise misclassifications.

The ground matrix of Wasserstein distance can be predefined with the experience to explore the pair-wise class correlation, $e.g.,$ the divergence of car and road is larger than car and bus. From the optimization perspective, we also set the ground metric to its increasing function. For semantic segmentation with unsupervised domain adaptation using constrained non-one-hot pseudo-label, we can also resort to the fast approximate solution of Wasserstein distance.

Instead of pre-defining the ground metric based on expert knowledge, we further propose to learn the optimal ground metric and a driving policy simultaneously in the CARLA simulator with an alternative optimization scheme. Our actor makes decision based on the latent representation of segmenter which is a partial observation of the front camera view. It can largely compress the state space for fast and stable training.

This paper is an extension of our preliminary segmentation work \cite{liuimportance,liu2020severity}. In summary, the contributions of this paper are summarized as

$\bullet$ We propose to render reliable segmentation results for autonomous driving by considering the different severity of misclassification. The inter-class severity is explicitly incorporated in the ground metric of our Wasserstein training framework. The importance-aware methods can be a particular case by designing a specific ground metric.

$\bullet$ The ground matrix can also be adaptively learned with an partially observable reinforcement learning framework based on the autonomous driving simulator with the alternative optimization.

$\bullet$ For both the one-hot and non-one-hot target label in self-training-based unsupervised domain adaption setting, we systematically explored the fast calculation for a non-negative linear, convex and concave function of ground metric.

We empirically validate its effectiveness and generality on multiple challenging benchmarks with different backbone models and achieve promising performance.

\section{Related Works}

\noindent\textbf{Semantic segmentation} predict a precise description of the class, location and shape \cite{badrinarayanan2017segnet}. The progress of deep learning \cite{liu2019hard,liu2018dependency,liu2018data,liu2018normalized,liu2018adaptive,liu2018joint,liu2017line} also contribute to a revolution semantic segmentation. \cite{long2015fully} developed a fully convolutional network for pixel-wise or superpixel-wise classification. The conventional methods usually adopt CE loss, which equally evaluates the errors incurred by all image pixels/classes without considering the different severity-level of different mistakes \cite{liuimportance,chen2018importance}.

The importance-aware methods \cite{chen2017importance,liuimportance} argue that the difference between object/pixel importance should be taken into account. The classes in Cityscapes are grouped as:

\noindent Group 4[most important]=$ \left\{{\rm Person, Car, Truck, Bus, \cdots}\right\}$;

\noindent Group 3=$ \left\{{\rm Road, Sidewalks, Train}\right\}$;

\noindent Group 2=$ \left\{{\rm Building, Wall, Fence, Vegetation, Terrain}\right\}$;

\noindent Group 1[least important]=$ \left\{{\rm Sky}\right\}$.

The more important group will be given larger weights to compute the sum of loss in all pixels. Therefore, the misclassification of a pixel with ground truth label in group 4 will result in a larger loss than misclassifying the sky to the other classes. However, its class-correlation is only defined in ground truth perspective rather than prediction classes. Recognizing a car to bus or road still receive the same loss is not sufficient for reliable autonomous driving. Besides, grouping manipulation is only based on human knowledge, which may differ from the way that machine perceives the world. Actually, this setting can be a special (but inferior) case of our framework.

Recently several powerful segmentation nets \cite{chen2017deeplab,romera2017erfnet} and the pose-processing strategies have also been developed to improve the initial results \cite{liu2015crf}. We note that this progress is orthogonal to our method and they can simply be added to each other. 

From the loss function perspective, the focal loss \cite{Lin_2017} is developed to balance the label distribution. \cite{li2017not} assign different pixel with different importance. \cite{berman2017lovszsoftmax} proposed a tractable surrogate for the optimization of the IoU measure. \cite{zhu2019improving} propose to improve the semantic segmentation performance via video propagation and label relaxation.

\noindent\textbf{Wasserstein distance} is a measure of distribution divergence \cite{kolouri2016sliced}. The Wasserstein distance or optimal transportation distance has attracted the attention of adversarial generative models \cite{arjovsky2017wasserstein}. However, the computing cost to solve the exact distance can be a large burden. Therefore, Wasserstein distance is usually hard to be used as the loss function. Several methods propose to approximate the Wasserstein distance, which has the complexity of $\mathcal{O}(N^2)$ \cite{cuturi2013sinkhorn}. \cite{frogner2015learning} propose to use it for the multi-class multi-label task with a linear model. Based on the previous Wasserstein loss works \cite{Han_2020_CVPR_Workshops,liu2019unimodal,liu2019conservative,liu2018ordinal}, we propose to adapt this idea to the severity-aware segmentation.

\noindent\textbf{Reinforcement learning (RL)} proposes to train an RL agent to play with a dynamic environment. The optimization objective of RL is to maximize its accumulated reward. The recent developed deep RL achieved the human-level performance in many Atari Games \cite{mnih2015human}.

End-to-end vision-based autonomous driving models \cite{dosovitskiy2017carla} trained by RL usually have a high computational cost. \cite{luck2016sparse} propose using variational inference to estimate policy parameters, while simultaneously uncovering a low dimensional latent space of actors. Similarly, \cite{haarnoja2018latent} analyze the utility of hierarchical representations for reuse in related tasks while learning latent space policies for RL. Recently, several works are proposed to combine semantic segmentation and policy learning
\cite{mousavian2019visual,muller2018driving,sax2019learning,zhou2019does}. We propose that the bottleneck of segmenter can be a natural representative lower-dimensional latent space which can efficiently shrink the state space and requires fewer actor parameters. Besides, we incorporate the RL in an alternative optimization framework to learn the optimal ground matrix in a simulator with a certain reward rule.

The CARLA simulator is a realistic environment for autonomous driving. Recently, many works are implemented on CARLA \cite{zhao2019lates,chen2019learning,toromanoff2019end,ohn2020learning,prakash2020exploring}. However, to the best of our knowledge, this is the first effort to define the inter-class correlations in CARLA.

\section{Methodology}

\begin{figure}[t]
\includegraphics[height=3.5cm]{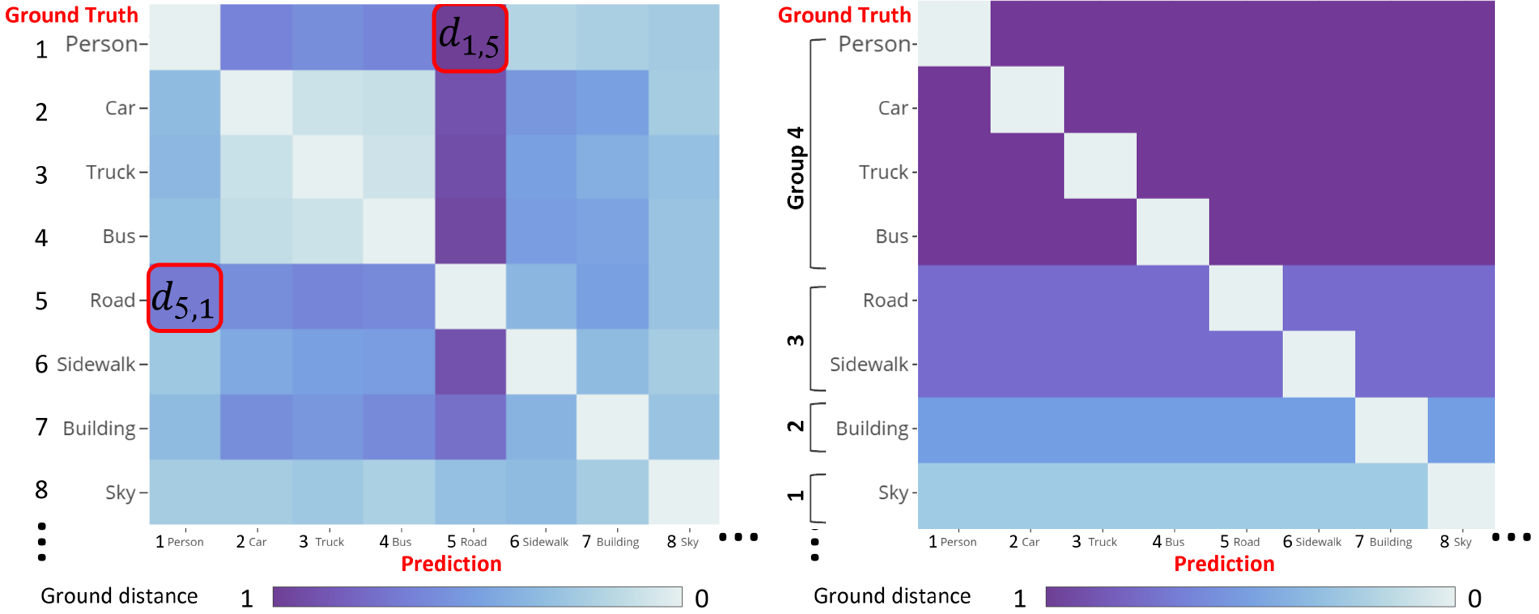}\\
\caption{Left: a possible ground matrix for severity-aware segmentation. Right: the ground matrix as an alternative for importance-aware setting.}
\label{fig:2}
\end{figure}

We target to learn a segmenter ${h}_\theta$, parameterized by $\theta$, with an autoencoder structure. It projects a street view image ${\rm\textbf{X}}\in\mathbb{R}^{H_x\times W_x\times 3}$ to a prediction of semantic segmentation map ${\rm\textbf{S}}\in\mathbb{R}^{H_s\times W_s\times N}$, where $N$ is the number of pre-defined classes in a segmentation dataset. We note the spatial size of input $H_x\times W_x$ and output $H_s\times W_s$ are not necessary the same or even have the shape of square in many segmenters. Let ${\rm\textbf{s}}=\left\{s_i\right\}_{i=1}^{N}$ be the prediction of a pixel in ${h}_\theta({\rm\textbf{X}})$, $i.e.,$ softmax normalized $N$ classes probability. $i\in\left\{1,\cdots,{\small N}\right\}$ be the index of dimension (class). We perform learning over a hypothesis space $\mathcal{H}$ of ${h}_\theta$. Given ${\rm\textbf{X}}$ and its target one-hot ground truth label ${\rm\textbf{T}}\in\mathbb{R}^{H_s\times W_s\times N}$, typically, learning is performed via empirical risk minimization to solve $\mathop{}_{{h}_\theta\in\mathcal{H}}^{\rm min}\mathcal{L}({h}_\theta({\rm\textbf{X}}),{\rm\textbf{T}})$, with a loss $\mathcal{L}(\cdot,\cdot)$ acting as a surrogate of performance measure. Following the previous segmentation works, we define the loss for each point. But we calculate the point-wise average of a mini-batch of images to update the networks.

Unfortunately, cross-entropy (CE)-based loss treat the output dimensions independently \cite{frogner2015learning}, ignoring the misclassification severity on label space.

Let us define ${\rm\textbf{t}}=\left\{t_j\right\}_{j=1}^{N}$ as the target histogram distribution label that can be either one-hot or non-one-hot vector. We assume the class label possesses a ground metric ${\rm\textbf{D}}_{i,j}$, which measures the severity of misclassifying $i$-th class pixel into $j$-th class. There are $N^2$ possible ${\rm\textbf{D}}_{i,j}$ in a $N$ class dataset and form a ground matrix $\textbf{D}\in\mathbb{R}^{N\times N}$ \cite{ruschendorf1985wasserstein}. When ${\rm\textbf{s}}$ and ${\rm\textbf{t}}$ are both histograms, the discrete measure of exact Wasserstein loss is defined as \begin{equation}
\mathcal{L}_{\textbf{D}_{i,j}}({\rm{\textbf{s},\textbf{t}}})=\mathop{}_{\textbf{W}}^{{\rm inf}}\sum_{j=0}^{N-1}\sum_{i=0}^{N-1}\textbf{D}_{i,j}\textbf{W}_{i,j} \label{con:df}
\end{equation} where \textbf{W} is the transportation matrix with \textbf{W}$_{i,j}$ indicating the mass moved from the $i^{th}$ point in source distribution to the $j^{th}$ target position. A valid transportation matrix \textbf{W} satisfies: $\textbf{W}_{i,j}\geq 0$; $\sum_{j=0}^{N-1}\textbf{W}_{i,j}\leq s_i$; $\sum_{i=0}^{N-1}\textbf{W}_{i,j}\leq t_j$; $\sum_{j=0}^{N-1}\sum_{i=0}^{N-1}\textbf{W}_{i,j}={\rm min}(\sum_{i=0}^{N-1}s_i,\sum_{j=0}^{N-1}t_j)$.

A possible ground matrix ${\rm\textbf{D}}$ in our application is shown in Fig. \ref{fig:2}. For instance, classifying the car to the road ($d_{2,5}$) has a larger ground metric than car to bus ($d_{2,4}$).

The Wasserstein distance is identical to the Earth mover's distance when the two distributions have the same total masses ($i.e., \sum_{i=0}^{N-1}s_i=\sum_{j=0}^{N-1}t_j$) and using the symmetric distance $d_{i,j}$ as ${\rm\textbf{D}}_{i,j}$. However, this is not true for our case. The entries in matrix ${\rm\textbf{D}}$ are not symmetric with respect to the main diagonal. For example, classifying the person to the road can be much severe than classifying the road to the person. Therefore, in Fig. \ref{fig:2}, $d_{1,4}$ should have a larger value than $d_{4,1}$. We note that the importance-aware learning can be achieved by configuring the ground matrix as Fig. \ref{fig:2}, which does not discriminate the different mistakes, $e.g.,$ classifying the car into any other classes has the same punishment. The groups also just pre-defined by human but not necessarily appropriate for practical driving system.

Actually, the simple version of IAL loss propose to assign a larger weight to the pixel position that has the ground truth label with more important level,  i.e., $j\ast\in$level 3 or 4. Considering that with the one-hot label encoding, the cross-entropy loss of each pixel is $-log s_{j\ast}$.


Since -log function is a deterministic function and is used for curving $s_{j\ast}$, the learning objective can be simplified as maximizing $w_{j\ast} s_{j\ast}$ to $w_{j\ast}$. Since $\sum s_j=1$, it is minimizing the $\sum_j w_{j\ast} s_{j}$ for $j\neq j\ast$. Our proposed Wasserstein loss can be $\sum_{i=0}^{N-1}f(d_{i,j\ast})s_i$. When we set $f(d_{i,j\ast})=w_{j\ast}$ for $i\neq j\ast$ and $f(d_{i,j\ast})=0$ for $i=j\ast$. The two losses are identical to each other.

\subsection{Wasserstein training with one-hot target}

The one-hot target vector ${\rm\textbf{t}}$ is the typical label for multi-class one-label dataset. We use $j$ to index the element of ${\rm\textbf{t}}$, and $j^\ast$ indicates the ground truth class. \footnote{\noindent We use $i,j$ interlaced for ${\rm \textbf{s}}$ and ${\rm \textbf{t}}$, since they index the same group of positions in a circle.}, and $0$ otherwise.

\noindent\textbf{Theorem 1.} \textit{Assume that} $\sum_{j=0}^{N-1}t_j=\sum_{i=0}^{N-1}s_i$, \textit{and} ${\rm{\textbf{t}}}$ \textit{is a one-hot distribution with} $t_{j^\ast}=1 ($or $\sum_{i=0}^{N-1}s_i)$\footnote{We note that softmax cannot strictly guarantee the sum of its outputs to be 1 considering the rounding operation. However, the difference of setting $t_{j^\ast}$ to $1$ or $\sum_{i=0}^{N-1}s_i)$ is not significant in our experiments using the typical format of softmax output which is accurate to 8 decimal places.}, \textit{there is only one feasible optimal transport plan.}

\begin{figure}[t]
\centering
\includegraphics[height=3.5cm]{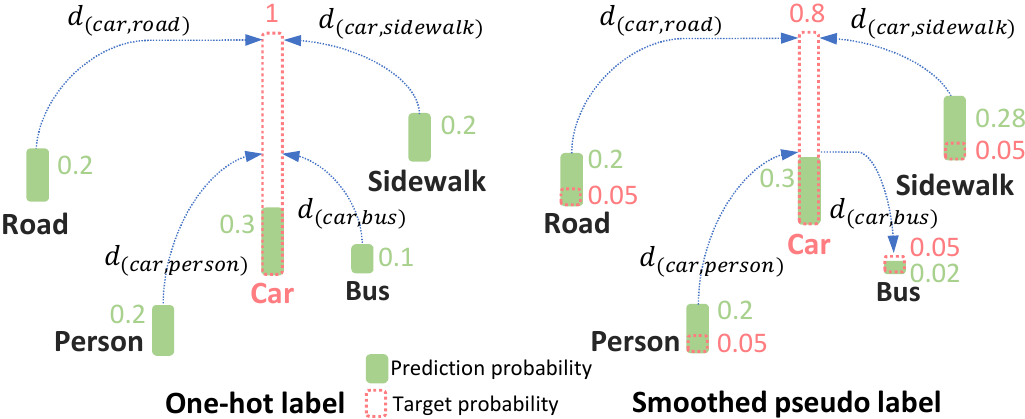}
\caption{Left: The only possible transport plan in one-hot target case. Right: the transportation in smoothed pseudo label ${\rm{\overline{\textbf{t}}}}$ is more complicated, $e.g.,$ car$\rightarrow$bus.}
\label{fig:3}
\end{figure}

According to the criteria of ${\rm\textbf{W}}$, all masses have to be transferred to the cluster of the ground truth label $j^\ast$, as illustrated in Fig. \ref{fig:3}. Then, the Wasserstein distance between softmax prediction {\rm{\textbf{s}}} and one-hot target {\rm{\textbf{t}}} degenerates to\begin{equation}
\mathcal{L}_{{\rm\textbf{D}}_{i,j}^{f}}({\rm{\textbf{s},\textbf{t}}})=\sum_{i=0}^{N-1} s_i f(d_{i,j^\ast}) \label{con:df}
\end{equation} We propose to extend the ground metric in ${\rm\textbf{D}}_{i,j}$ as $f(d_{i,j})$, where $f$ can be a linear or increasing function proper, $e.g., p^{th}$ power of $d_{i,j}$ and Huber function. The exact solution of Eq. \eqref{con:df} can be computed with a complexity of $\mathcal{O}(N)$. The ground metric term $f(d_{i,j^\ast})$ works as the weights $w.r.t.$ $s_i$, which takes all classes into account following a soft attention scheme \cite{liu2018dependency}. It explicitly encourages the probabilities distributing on the neighboring classes of $j^\ast$.

In contrast, the CE loss in one-hot setting can be formulated as $-1{\rm log}s_{j^\ast}$. Similar to the hard prediction scheme, only a single class prediction is considered resulting in a large information loss \cite{liu2018dependency}. Besides, the regression loss with softmax prediction could be $f(d_{i^\ast,j^\ast})$, where $i^\ast$ is the class with maximum prediction probability.

\subsection{Monotonic increasing $f$ w.r.t. $d_{i,j}$ as ground metric}

Practically, $f$ in ${\rm\textbf{D}}_{i,j}^f=f(d_{i,j})$  can be a positive increasing mapping function $w.r.t.$ $d_{i,j}$ for better optimization. Although the linear function is satisfactory for comparing the similarity of SIFT or hue \cite{rubner2000earth}, which do not involve neural network optimization.

\noindent $\bullet$ \textbf{Convex function $ w.r.t.$ $d_{i,j}$ as the ground metric.} We can extend the ground metric as a nonnegative increasing and convex function of $d_{i,j}$. Here, we give some measures\footnote{We refer to ``measure'', since a $\rho^{th}$-root normalization is required to get a distance \cite{villani2003topics}, which satisfies three properties: positive definiteness, symmetry and triangle inequality.} using the typical convex ground metric function.

$\mathcal{L}_{{\rm\textbf{D}}_{i,j}^\rho}{(\rm{{\textbf{s},{\textbf{t}}}})}$, the Wasserstein measure using $d^\rho$ as the ground metric with $\rho=2,3,\cdots$. The case $\rho=2$ is equivalent to the Cram\'{e}r distance \cite{rizzo2016energy}. Note that the Cram\'{e}r distance is not a distance metric proper. However, its square root is.\begin{equation}
{\rm\textbf{D}}_{i,j}^\rho= d_{i,j}^\rho    
\end{equation}

\vspace{-3pt}
$\mathcal{L}_{{\rm\textbf{D}}_{i,j}^{H\tau}}{(\rm{{\textbf{s},{\textbf{t}}}})}$, the Wasserstein measure using a Huber cost function with a parameter $\tau$.\begin{equation}
{\rm\textbf{D}}_{i,j}^{H\tau}=\left\{
             \begin{array}{ll}
             d_{i,j}^2&{\rm{if}}~d_{i,j}\leq\tau\\
             \tau(2d_{i,j}-\tau)&{\rm{otherwise}}.\\
             \end{array}
             \right.
\end{equation}

\noindent $\bullet$ \textbf{Concave function $ w.r.t.$ $d_{i,j}$ as the ground metric.} In practice, it may be not meaningful to set the ground metric as a nonnegative, concave and increasing function $w.r.t.$ $d_{i,j}$. 

We note that the computation speed of exact solution in conservative target label case is usually not satisfactory, but the step function $f(t)=\mathbbm{1}_{t\neq0}$ (one everywhere except at 0) can be a special case, which has significantly less complexity \cite{villani2003topics}. Assuming that the $f(t)=\mathbbm{1}_{t\neq 0}$, the Wasserstein metric between two normalized discrete histograms on $N$ bins is simplified to the $\ell_1$ distance. \begin{equation}
\mathcal{L}_{\mathbbm{1}{d_{i,j}\neq 0}}{(\rm{{\textbf{s},{\textbf{t}}}})}=\frac{1}{2}\sum_{i=0}^{N-1}{|{\rm{s}}_i-{\rm{{t}}}_i|}=\frac{1}{2}||{\rm{\textbf{s}}}-{\rm{{\textbf{t}}}}||_1
\end{equation} where $||\cdot||_1$ is the discrete $\ell_1$ norm. Unfortunately, its fast computation is at the cost of losing the ability to discriminate the difference of probability in different bins.

\subsection{Learn severity-aware ground matrix}

Other than the pre-defined ground matrix, we further propose to learn the ground matrix in a simulator with our autonomous driving agent following the alternative optimization.

\begin{figure}[t]
\centering
\includegraphics[height=5cm]{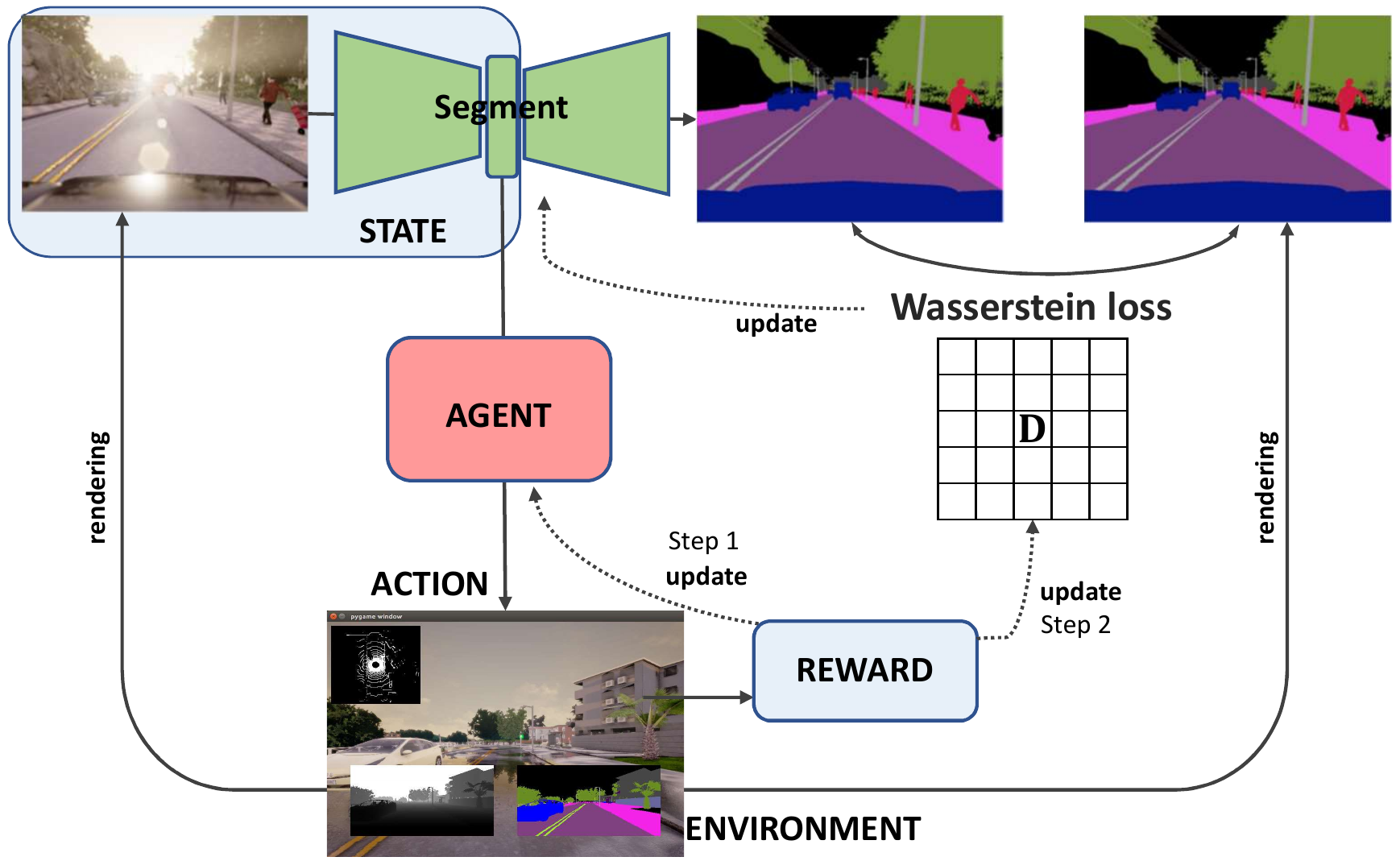}
\caption{The reinforced alternative optimization framework to learn actor-critic agent and ground matrix simultaneously.}
\label{fig:4}
\end{figure}

The overall framework is illustrated in Fig. \ref{fig:4}. We choose a high-reality simulator, the CARLA \cite{dosovitskiy2017carla}, as our environment. The view of a monocular camera placed at the front the car is rendered as $\textbf{X}$. Segmenter takes $\textbf{X}$ as input and predicts the segmentation image $\textbf{S}$ which is compared with target $\textbf{T}$ with Wasserstein loss.

An agent learns to interact with the environment following a partially observable Markov decision process (POMDP). For the time step $t$, a RL agent observes the state ${s_t}$ in a state space $\mathcal{S}$ and predict an action ${a_t}$ from an action space $\mathcal{A}$, following the RL policy $\pi(a_t|s_t)$, which is the behavior of the agent. Then, the action will result in the change of environment and move to the next state $s_{t+1}$, and receive a reward ${r_t(s_t,a_t)}\in \mathcal{R} \subseteq \mathbb{R}$ from the dynamic environment. The optimization objective of an optimal policy $\pi^\ast$ is to maximize the discounted total return $R_t=\sum_{i\geq0}^{T}\gamma^{i}{r_{t+i}(s_t,a_t)}$ in expectation, where $\gamma\in [0,1)$ is used to balance the current and the long-term rewards \cite{li2017deep}.

Instead of using $\textbf{X}$ as our state \cite{dosovitskiy2017carla}, we propose to utilize the latent representation of our segmenter. It can be either feature vector or feature maps according to the backbone. \cite{dosovitskiy2017carla} takes 12 days for the training on CARLA with only 84$\times$84 size raw image. As a partial observation, the latent representation compresses the state space drastically. Compared to the raw image, segmentation map or its latent representation has sufficient information ($e.g.,$ each object and their and precise location) to guide the driving, and is robust to appearance variation ($e.g.,$ weather, lighting). Since a high proportion of pixels have the same label as their neighbors in $\textbf{S}$, there are a large of room to reduce its redundancy.

The network takes two latent representations as input, which is the two most recent at this step, as well as a vector of sensor readings. The two inputs are feed to two different branches: feature maps by a convolutional module, measurements by a fully-connected network. The two branches are merged later and further process the fused information.

In the context of autonomous driving, we define the action as a three dimensional vector for steering ${a^s_t}\in[-1,1]$, throttle ${a^t_t}\in[0,1]$ and brake ${a^b_t}\in[0,1]$. We define the reward ${r_t}=1-\alpha o_l-\beta o_r- \psi c$, where $o_l,o_r\in[0,1]$ measure the degree of off-line or off-road respectively, and $c\in\left\{0, \frac{1}{4}, \frac{1}{2}, \frac{3}{4}, 1\right\}$ indicates there is no/S0/S1/S2/S3 level crash, where S0, S1, S2, S3 denotes the severity is negligible/minor, major, hazardous and catastrophic defined in [ISO26262] \cite{iso201126262}. $\alpha,\beta$ and $\psi$ are a set of positive weights to balance the punishments, we empirically set $\alpha=1,\beta=1$ and $\psi=10$ in all of our experiences. The agent will receive the reward of 1 when the vehicle drives smoothly and keep in line and road. The driving will be terminated when there is a crash / completely (100\%) off-line / 50\% off-road / reaches 500 time steps.

Given a continuous action space, the value-based RL, for example the Q-Learning, cannot be able to pedict continuous values. Therefore, we resort to the actor-critic algorithm. As a kind of the policy-based method, the objective of RL is to learn a policy ${\pi}_\theta(a_t|s_t)$ to maximize the expected reward $J(\theta)$ over all possible decisions. With the policy gradient theorem \cite{sutton2000policy}, the gradient of the parameters given the objective function has the form: \begin{align}
 {\nabla}_\theta J(\theta)=\mathbb{E}[{\nabla}_\theta \text{log} {\pi}_\theta(a_t|s_t)({Q}(s_t,a_t)-b(s_t))] 
\end{align}

\noindent where ${Q}(s_t,a_t)=\mathbb{E}[R_t|s_t,a_t]$ can be defined as the state-action value function. We note that the initial action $a_t$ is provided to calculate the expected return when starting in the state $s_t$. Moreover, the baseline function $b(s_t)$ is usually subtracted to reduce the variance and not changing the estimated gradient \cite{williams1992simple,andrew1999reinforcement}. A possible baseline function can be the state only value function ${V}(s_t)=\mathbb{E}[R_t|s_t]$. It is similar to $Q(s_t,a_t)$, except the $a_t$ is not given here. The advantage function is defined as ${A}(s_t,a_t)={{Q}(s_t,a_t)}-{{V}(s_t)}$ \cite{li2017deep}. Eq.(4) then becomes:\begin{align}
 {\nabla}_\theta J(\theta)=\mathbb{E}[{\nabla}_\theta \text{log} {\pi}_\theta(a_t|s_t){A}(s_t,a_t)] 
\end{align} 

It can be regarded as a specific case of actor-critic RL, in which ${\pi}_\theta(a_t|s_t)$ can be the actor and the ${A}(s_t,a_t)$ can be the critic. To reduce the number of required parameters, the parameterized temporal difference error ${\delta_\omega}={r_t}+{\gamma}V_{\omega}(S_{s+1})-V_{\omega}(S_{s})$ can be used to approximate the advantage function. We adopt two different symbols $\theta$ and $\omega$ to denote the actor and critic function respectively. We note that the most of these parameters are shared in a mainstream neural network, then separated to two branches for policy and value predictions. We further adapt the A3C to its off-policy version to stabilize and speed up our training.

After configuring our RL module, we propose to adaptively learn the ground metric along with the training of actor following the alternative optimization.

\noindent\textbf{Step 1:} Fixing the ground matrix to compute $\mathcal{L}_{{\rm\textbf{D}}_{i,j}}({\rm{\textbf{s},\textbf{t}}})$ and updating the network parameters of our actor-critic module.

\noindent\textbf{Step 2:} Fixing the network parameters and postprocessing the ground matrix with the feature-level $\ell_1$ distances between different classes.

In this round, we use the normalized second-to-last convolutional layer's channel-wise response at each point as a feature vector, since there is no subsequent non-linearities. Therefore, it is meaningful to average the feature vectors in each position that corresponds to the pixel in image-level with the same class label to compute their centroid and reconstruct ${{\rm\textbf{D}}_{i,j}}$ using the $\ell_1$ distances between these centroids $\overline{d}_{i,j}$. To avoid the model collapse, we construct the ${{\rm\textbf{D}}_{i,j}=\frac{1}{1+\alpha}\left\{f(\overline{d}_{i,j})+\alpha f(d_{i,j})\right\}}$ in each round, and decrease $\alpha$ from 10 to 0 gradually in the training.

\section{Experiments}

In the experiment section, we provide the implementation details and experimental results on two typical autonomous driving benchmarks ($i.e.,$ Cityscapes \cite{cordts2016cityscapes} and CamVid \cite{brostow2009semantic}) and the CARLA simulator \cite{dosovitskiy2017carla}. Other than the comparisons, We also give the detailed ablation study to illustrate the effectiveness of each module and their combinations. Our Wasserstein loss framework is implemented in PyTorch platform. All of the networks are pre-trained with CE loss as their vanilla version.

We follow the RL agent structure proposed in \cite{volodymyr2015human}. The two most recent latent feature maps observed by the agent and a vector of measurements are feed into the two branches of the agent. The measurement vector includes the current speed of the car, distance to the goal, damage from collisions, and the current high-level command provided by the topological planner, in one-hot encoding. We note that the inputs are processed by two separate branches. Specifically, the feature maps is feed to a convolutional branch, while the measurements are feed to a fully-connected branch. After the processing, we concatenate the two outputs to fuse the information.

Our RL framework is trained with 10 parallel actor threads, for a total of 10 million environment steps. As in previous work \cite{jaderberg2016reinforcement}, we also choose 20-step rollouts. The initial learning rate is set to 0.0007, and with the entropy regularization of 0.01. Along with the training, the learning rate our network is linearly decreased to zero. We note that the Wasserstein loss is defined for each pixel in the image, but we calculate the point-wise average of a mini-batch of images to update the networks.

\begin{figure}[t]
\centering
\includegraphics[height=7cm]{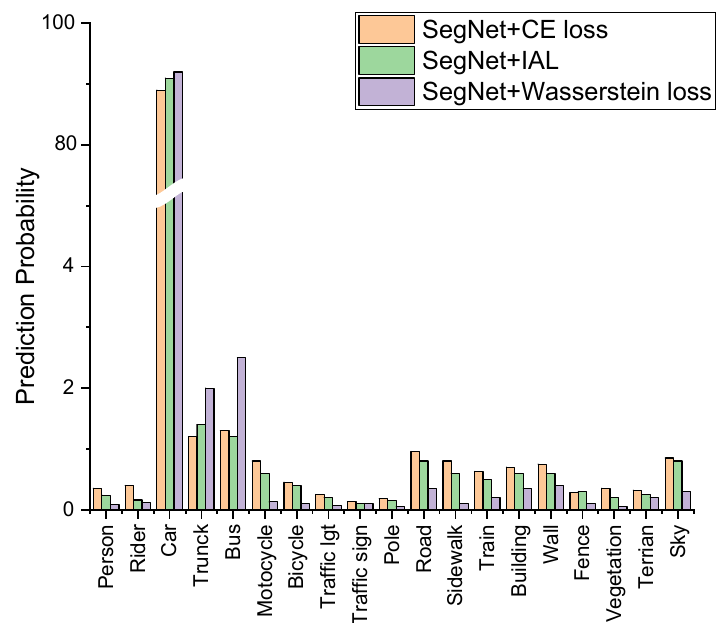}
\caption{The confusion statistics of classifying car on the testing set of Cityscapes dataset with SegNet backbone.}
\label{fig:s2}
\end{figure}

According to CARLA simulator \cite{dosovitskiy2017carla}, the inputs are the camera image and the sensor reading information. We adopt two fully connected (FC) layers (64,64) to process the vector of sensor reading. We apply two convolutional layers with 3$\times$3$\times$32 and 3$\times$3$\times$16 kernels and followed by two fully connected layers (1024,512). Since the latent feature map of different segmentation backbone has a different size, the trained network of this part cannot be shared among different backbones. As shown in Fig. \ref{fig:s2}, our actor-critic uses two fully connected (FC) layers (256,128) then cascade two sub-branches with two fully connected layers (64,16). The number of the output unit is set as 3 which indicates the steering, throttle and brake.

The evaluation using third-party reinforcement framework on CARLA follows the experiment setting, network structures and hyperparameter settings as \cite{fennessy2019autonomous}\footnote{\url{https://gitlab.com/grant.fennessy/rl-carla}}.

\begin{table}[t]  
\resizebox{\linewidth}{!}{
\centering
\begin{tabular}{|l|c|c|c|c|c|c|c|c|}
\hline
&\multicolumn{7}{c|}{Group4}&\multirow{2}{*}{mIoU}\\\cline{2-8}

&Person&Rider&Car&Truck&Bus&Motor&Bike&\\\hline\hline
SegNet&62.8&42.8&89.3&38.1&43.1&35.8&51.9&57.0\\\hline
+IAL&84.1&46.0&91.1&75.9&65.0&22.2&\textbf{65.3}&{65.7}\\\hline
+$\mathcal{L}_{d_{i,j}}$&86.4&48.7&92.8&78.5&68.2&40.2&62.8&67.4\\\hline 
+$\mathcal{L}_{{\rm\textbf{D}}_{i,j}^2}$&87.5&\textbf{50.2}&\textbf{93.4}&\textbf{79.8}&69.5&\textbf{42.0}&64.3&\textbf{68.0}\\\hline
+$\mathcal{L}_{{\rm\textbf{D}}_{i,j}^{H\tau}}$&\textbf{87.6}&49.8&93.2&79.5&\textbf{70.3}&41.6&63.6&67.9\\\hline
+$\mathcal{L}_{\mathbbm{1}}$&63.0&41.5&87.4&40.1&43.7&38.2&50.6&56.3\\\hline\hline
ENet&65.5&38.4&90.6&36.9&50.5&38.8&55.4&58.3\\\hline
+IAL&87.7&41.3&92.4&\textbf{73.5}&76.2&24.1&69.7&67.5\\\hline
+$\mathcal{L}_{d_{i,j}}$&90.7&48.7&95.5&70.8&75.3&46.2&73.3&69.1\\\hline 
+$\mathcal{L}_{{\rm\textbf{D}}_{i,j}^2}$&90.9&\textbf{49.6}&\textbf{96.8}&71.4&77.6&\textbf{46.3}&\textbf{75.1}&69.3\\\hline
+$\mathcal{L}_{{\rm\textbf{D}}_{i,j}^{H\tau}}$&\textbf{90.1}&49.5&\textbf{96.8}&72.6&\textbf{77.8}&46.2&75.0&\textbf{69.5}\\\hline
+$\mathcal{L}_{\mathbbm{1}}$&72.5&40.3&85.2&39.4&48.7&41.0&52.9&59.1\\\hline\hline 
FCN &75.4& 50.5& 91.9& 35.3 &49.1& 50.7 &65.2 &64.3\\\hline 
+IAL &90.4& 56.6 &93.7 &68.5 &74.6& 31.5 &\textbf{81.5}& 71.9\\\hline 
+$\mathcal{L}_{d_{i,j}}$&89.5&60.3&92.5&73.2&73.5&54.2&71.0&71.7\\\hline 
+$\mathcal{L}_{{\rm\textbf{D}}_{i,j}^2}$&90.6&{56.5}&{93.8}&\textbf{74.6}&74.4&\textbf{56.1}&{70.3}&72.0\\\hline
+$\mathcal{L}_{{\rm\textbf{D}}_{i,j}^{H\tau}}$&\textbf{91.5}&\textbf{59.4}&\textbf{95.2}&{74.3}&\textbf{74.6}&52.4&72.4&\textbf{72.2}\\\hline
+$\mathcal{L}_{\mathbbm{1}}$&78.3&60.1&88.4&49.5&52.2&51.6&69.1&65.2\\\hline
\end{tabular}
}
\caption{The comparison results of various methods of Cityscapes Group 4 with SegNet, ENet and FCN backbone.}\label{tab:1}
\end{table}

We introduce the used evaluation metrics as follow.

\noindent $\bullet$ The intersection-over-union (IoU) is defined as:

\begin{equation}
{\rm IoU}=\frac{{\rm TP}}{\rm TP+FP+FN}
\end{equation} where TP, FP, and FN denote the numbers of true positive, false positive, and false negative pixels, respectively. Moreover, the mean IoU is the average of IoU among all classes.\\

\noindent$\bullet$ {Metrics used in third-party evaluation \cite{fennessy2019autonomous}}

The metrics used in the third-party evaluation in CARLA ($i.e.$, Table \textcolor{red}{5}) are as follows:

\noindent\textbf{Drive\%} measures the number of steps that took place during the evaluation divided by 720,000. A value of 100\% indicates the agent does not have the early termination. In contrast, a low value indicates the failures.

\noindent\textbf{Km}is the total kilometers driven across all steps. It is also the function of mean speed and drive\%.

\begin{table}[t]
\centering
\resizebox{1\linewidth}{!}{%
\begin{tabular}{|l|c|c|c|c|c|c|c|}
\hline
&\multicolumn{3}{c|}{Group3}&\multicolumn{3}{c|}{Group4}&\multirow{2}{*}{mIoU}\\\cline{2-7}
&Road&Sidewalk&Sign&Car&Pedestrian&Bike&\\\hline\hline
FCN&98.1&89.5&25.1&84.5&64.6&38.6&69.6\\\hline
+IAL&96.3&91.8&21.5&82.2&69.5&57.6&71.2\\\hline
+$\mathcal{L}_{d_{i,j}}$&98.5&93.2&28.3&87.4&71.3&60.0&72.4\\\hline 
+$\mathcal{L}_{{\rm\textbf{D}}_{i,j}^2}$&\textbf{98.7}&94.6&\textbf{29.7}&89.5&73.4&\textbf{60.7}&\textbf{72.8}\\\hline
+$\mathcal{L}_{{\rm\textbf{D}}_{i,j}^{H\tau}}$&98.5&\textbf{95.0}&29.5&\textbf{89.7}&\textbf{73.5}&60.6&\textbf{72.8}\\\hline
\end{tabular}}
\caption{The comparison results of various methods on the Group 3/4 of CamVid dataset using FCN as backbone.}\label{tab:2}
\end{table}

\noindent\textbf{Km/Hr} denotes the mean speed taken across all steps. The pre-set maximum speed in CARLA is 25km/hr.

\noindent\textbf{Km/OOL} denotes the driving distance on average between each out of lane (OOL) instances. Ideally, this value is infinite if there is no OOL infraction, the value can be infinite.

\noindent\textbf{Km/Collision} denotes the driving distance on average between each collision with an object in the environment. Ideally, this value is infinite if no collisions occur.

\begin{table}[t!]
\resizebox{\linewidth}{!}{
\centering
\begin{tabular}{|c|c|c|c|c|c|c|c|c|}
\hline
&\multicolumn{7}{c|}{Group4}&\multirow{2}{*}{mIoU}\\\cline{2-8}
&Person&Rider&Car&Truck&Bus&Motor&Bike&\\\hline\hline
LRENT&61.7&27.4&83.5&27.3&37.8&30.9&41.1&46.5\\\hline
$\mathcal{L}_{d_{i,j}}$&65.4&33.7&88.5&36.2&44.8&39.3&48.4&47.8\\\hline 
$\mathcal{L}_{{\rm\textbf{D}}_{i,j}^2}$&65.7&34.0&88.9&36.7&45.3&39.6&49.1&48.0\\\hline
$\mathcal{L}_{{\rm\textbf{D}}_{i,j}^{H\tau}}$&\textbf{66.2}&\textbf{34.7}&\textbf{89.5}&\textbf{37.1}&\textbf{46.0}&\textbf{40.8}&\textbf{50.5}&\textbf{48.3}\\\hline
\end{tabular}
}
\caption{The comparison results of various methods on the Group4 of GTA5$\rightarrow$Cityscapes unsupervised domain adaptation using DeeplabV2 as backbone.}\label{tab:3}
\end{table}

\subsection{Importance-aware SS with one-hot label}
We firstly pre-define our ground matrix as Fig. \ref{fig:2} right to achieve the importance-aware SS. Following the setting in IAL \cite{chen2017importance,chen2018importance}, we choose the SegNet \cite{badrinarayanan2017segnet} and ENet \cite{paszke2016enet} as our backbone to fairly compair with IAL. We note that our method can be applied on more advanced backbone \cite{romera2017erfnet}. The conventional CE loss in their vanilla version is replaced by IAL and our Wasserstein loss.

The recent Cityscapes dataset has 2975/500/1525 images for training/validation/testing respectively. The 19 most frequently used classes are chosen and grouped as IAL. Table \ref{tab:1} shows that the class in group 4 are segmented with higher IoU when considering the importance of each class. Our Wasserstein loss usually outperforms IAL by more than 2\%, especially apply the convex function $w.r.t. d_{i,j}$. The improvements $w.r.t.$ Motor are more than 15\% over IAL.

The CamVid dataset contains 367, 26 and 233 images for training, validation, and testing respectively. To make fair evaluation, we choose the same setting and measurements as IAL, and report the results in Table \ref{tab:2}. We note that fine-tuning a public available trained FCN segmenter \cite{long2015fully} with Wasserstein loss is 1.5$\times$ faster than the training of IAL. We note that the training use only Wasserstein loss can be 2.2 or 2.4 times slower than CE loss in Cityscapes or CamVid datasets respectively. We have added the related comparison in our revised version. Although the IoU of some unimportant classes may drop, this will have a limited impact on driving safety. We note that the mean IoU of all classes can still be comparable or improved since we introduced a more strict objective than CE loss only. Since the metrics used in IAL cannot evidence the superiority of severity-aware setting, we give additional confusion statistics in figure \ref{fig:s2}.

We can see that the prediction probability of SegNet+Wasserstein training is more concentrate to car/truck/bus. Although the improvement of correctly classifying car as car is about 1\% to 3\% over IAL or SegNet as shown in Table 1, IAL/SegNet has more severe misclassifications, e.g., car$\rightarrow$person and/rider/motor/bike/sky. Noticing that our correct classification probabilities in other classes are usually more significant and promising than car, we just pick one that has similar correct probability class and show how different they make mistakes. Even they have similar probability to be wrong, their consequences will have different severity.

\begin{figure}[t]
\centering
\includegraphics[width=8cm]{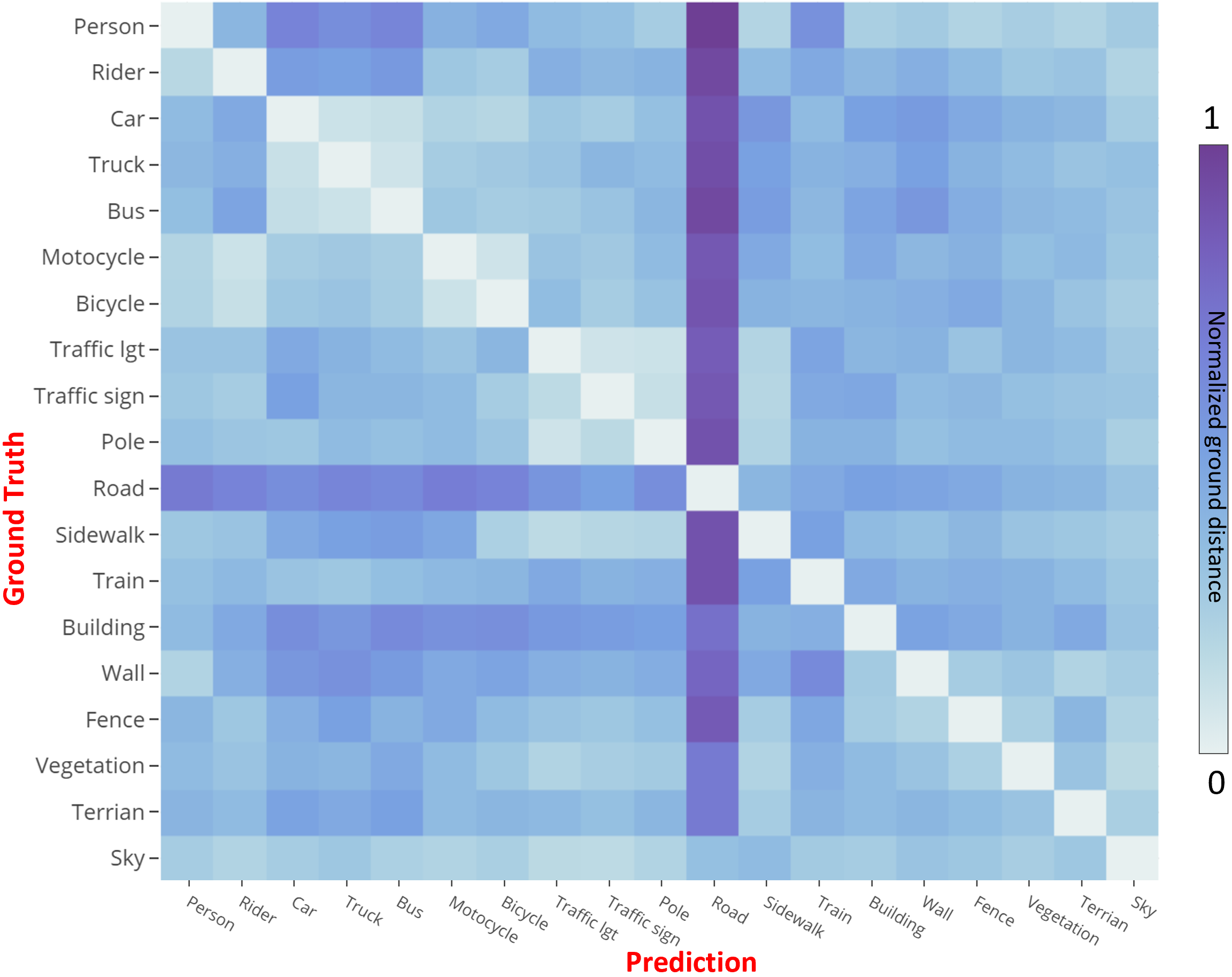}\\
\caption{Normalized adaptively learned ground matrix in CARLA simulator with ENet backbone.}\label{fig:5}
\end{figure}

\subsection{Wasserstein training with conservative target}

The self-training scheme \cite{zou2019confidence} can be a promising solution for the unsupervised domain adaptation in both classification and semantic segmentation \cite{romera2019bridging}, which involves an iterative process. Specifically, it first predict on the target domain and then taking the confident predictions as pseudo-labels for retraining. Unavoidable, the pseudo-labels can be noisy and unreliable. In consequence, the self-training can put overconfident label belief on wrong classes, leading to deviated solutions with propagated errors. \cite{zou2019confidence} propose to construct the smoothed pseudo-label ${\rm{\overline{\textbf{t}}}}$, which smooth the one-hot pseudo-label to a conservative (i.e., non one-hot) target distribution. Using the conservative distribution as the label, the fast computing of Wasserstein distance in Eq. \eqref{con:df} is not applicable.

The closed-form result of the general Wasserstein distance can have the complexity higher than $\mathcal{O}(N^3)$, which cannot satisfy the speed requirement of the loss function. Therefore, a possible solution is to approximate the Wasserstein distance, which usually has the complexity of $\mathcal{O}(N^2)$. \cite{cuturi2013sinkhorn} proposes an efficient approximation of both the transport matrix in and the subgradient of the loss, which is essentially a matrix balancing problem that has been well-studied in numerical linear algebra \cite{knight2013fast}.

\subsection{Importance-aware SS with conservative label}

We further test our method for unsupervised domain adaptation with constrained self-training, i.e., label entropy regularizer (LRENT) \cite{zou2019confidence}. We compute the approximate Wasserstein distance as the loss. Table \ref{tab:3} shows the performance of GTA5$\rightarrow$Cityscapes adaptation and outperforms the CE loss-based LRENT by more than 5\% in these important classes consistently. The improvements of $\mathcal{L}_{{\rm\textbf{D}}_{i,j}^{H\tau}}$over $\mathcal{L}_{{\rm\textbf{D}}_{i,j}^2}$ are more significant than the one-hot case. This is probably because that the Huber function is more robust to the label noise which is common for the pseudo label in self-learning method. This task also indicates that our method can be a general alternative objective of CE loss and be applied in a plug and play fashion.  We note that using Eq. 6, the Wasserstein loss will totally lost the discriminability of different missclassification.

\begin{figure}[t]
\begin{center}
\resizebox{1\textwidth}{!}{
\begin{tabular}{@{}cccccccccc@{}}
\cellcolor{city_color_1}\textcolor{white}{~~road~~} &
\cellcolor{city_color_2}~~sidewalk~~&
\cellcolor{city_color_3}\textcolor{white}{~~building~~} &
\cellcolor{city_color_4}\textcolor{white}{~~wall~~} &
\cellcolor{city_color_5}~~fence~~ &
\cellcolor{city_color_6}~~pole~~ &
\cellcolor{city_color_7}~~traffic lgt~~ &
\cellcolor{city_color_8}~~traffic sgn~~ &
\cellcolor{city_color_9}~~vegetation~~ & 
\cellcolor{city_color_0}\textcolor{white}{~~ignored~~}\\
\cellcolor{city_color_10}~~terrain~~ &
\cellcolor{city_color_11}~~sky~~ &
\cellcolor{city_color_12}\textcolor{white}{~~person~~} &
\cellcolor{city_color_13}\textcolor{white}{~~rider~~} &
\cellcolor{city_color_14}\textcolor{white}{~~car~~} &
\cellcolor{city_color_15}\textcolor{white}{~~truck~~} &
\cellcolor{city_color_16}\textcolor{white}{~~bus~~} &
\cellcolor{city_color_17}\textcolor{white}{~~train~~} &
\cellcolor{city_color_18}\textcolor{white}{~~motorcycle~~} &
\cellcolor{city_color_19}\textcolor{white}{~~bike~~}
\end{tabular}
}
\end{center}
\includegraphics[width=8.2cm]{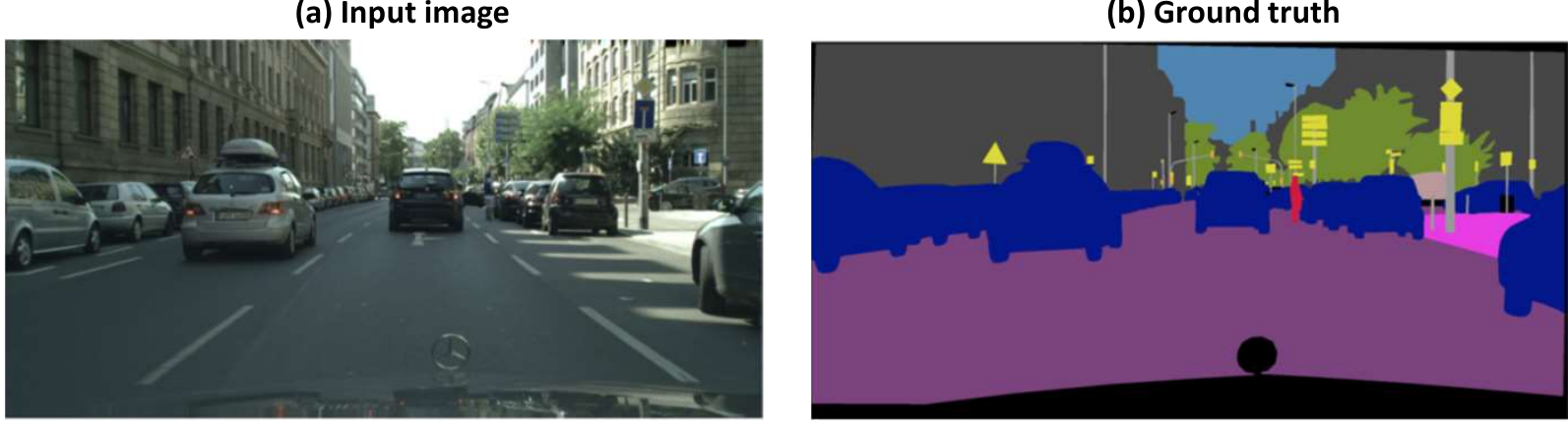}\\

\includegraphics[width=8.2cm]{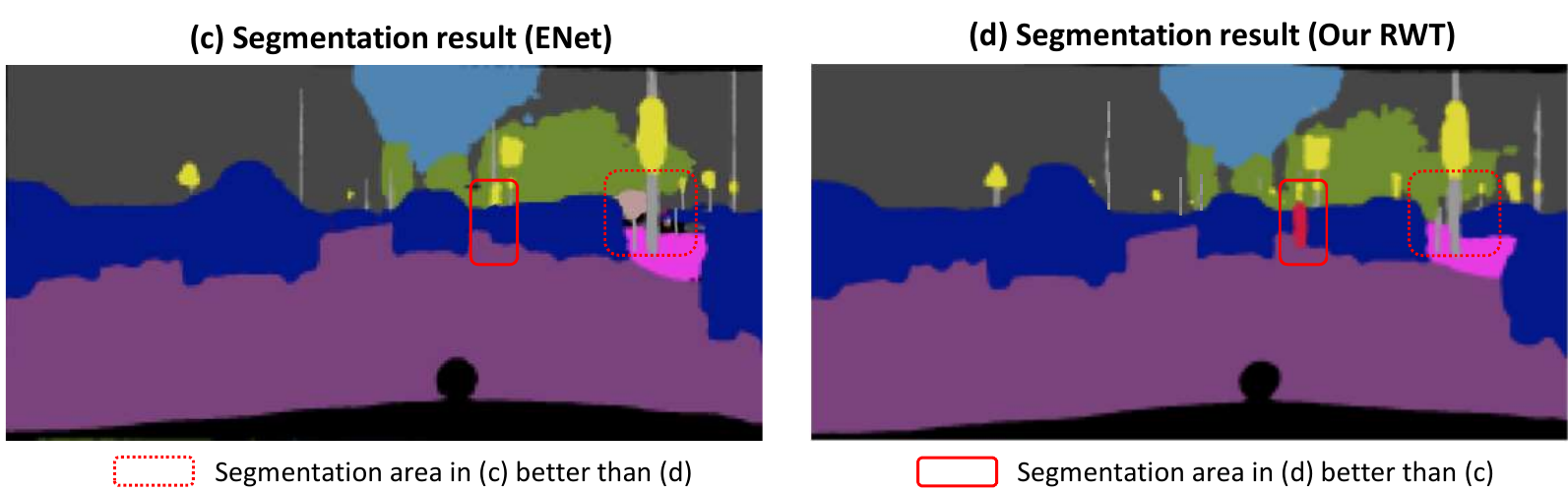}

\caption{Representative semantic segmentation result of ENet and our reinforcement Wasserstein training with ENet backbone on Cityscapes dataset. The two image has the same mIoU but the misclassification of the person may lead to more severity result.}
\label{fig:6}
\end{figure}

\subsection{Severity-aware SS with learned ground matrix}

As discussed in our introduction, the importance-aware setting does not consider the different severity $w.r.t.$ the predictions. Instead of pre-define a severity-aware ground matrix with human knowledge, we propose to learn it in the CARLA simulator\footnote{\url{https://carla.org}} and show our result with ENet backbone in Fig. \ref{fig:5}. We train our actor-critic with 10 parallel actor threads as \cite{dosovitskiy2017carla} for a total of 5-million steps. The joint learning of our actor-critic module and the ground matrix only takes 10.5 hours which is much faster than using the images as the state. We note that \cite{dosovitskiy2017carla} takes 12 days to train a reinforcement learning framework. The time cost will be intractable when we incorporate a ground matrix simultaneously. To evidence the effectiveness of our method, we show a segmentation example in Fig. \ref{fig:6}.

\begin{table}[t]  
\resizebox{\linewidth}{!}{
\centering
\begin{tabular}{|c|c|c|c|c|c|c|}
\hline
\multirow{2}{*}{Task}&\multicolumn{2}{c|}{Training condition}&\multicolumn{2}{c|}{New town}&\multicolumn{2}{c|}{New weather}\\\cline{2-7}

&wo/&w/&wo/&w/&wo/&w/\\\hline\hline

collision-person&12.61&\textbf{30.43}&2.53&\textbf{7.82}&9.24&\textbf{28.25}\\\hline
collision-car&   0.84&\textbf{4.59}&0.40&\textbf{2.79}&0.75&\textbf{4.33}\\\hline 
collision-static&0.45&\textbf{1.36}&0.26&\textbf{1.02}&0.28&\textbf{1.29}\\\hline
off-line&        0.18&\textbf{0.85}&0.21&\textbf{0.78}&0.14&\textbf{0.81}\\\hline
off-road&        0.76&\textbf{1.47}&0.43&\textbf{1.22}&0.71&\textbf{1.35}\\\hline 
\end{tabular}
}
\caption{The average distance (km) between the two infractions of using the ENet trained only with CE loss (wo/) or fine-tuned with Wasserstein loss (w/) in our reinforcement learning framework. Higher is better.}\label{tab:4}
\end{table}

\begin{table}[t!]  
\resizebox{\linewidth}{!}{
\centering
\begin{tabular}{|c|c|c|c|c|c|c|}
\hline
Method&Drive$\%$&Km&Km/Hr&Km/Off-line&Km/Collision\\\hline\hline 
Deeplab wo/&82.2&31.9&9.3&0.04&12.4\\\hline
Deeplab w/ IAL&85.8&35.2&12.4&0.08&15.7\\\hline 
Deeplab w/ A-${\mathcal{L}_{d_{i,j}}}$&\textbf{91.6}&\textbf{47.5}&\textbf{20.4}&\textbf{0.14}&\textbf{20.7}\\\hline 
\end{tabular}
}
\caption{Results of different training methods using Deeplab backbone and Deeplab/\cite{fennessy2019autonomous} evaluation on the CARLA simulator. Higher is better.}\label{tab:5}
\end{table}

\begin{table}[t!]  
\resizebox{\linewidth}{!}{
\centering
\begin{tabular}{|c|c|c|c|}
\hline
Method&   Training & New town & New Weather \\\hline\hline 
Deeplab wo/&  58.2 &  33.7  & 30.5 \\\hline
Deeplab w/ IAL&  62.5 & 38.3   & 35.2  \\\hline 
Deeplab w/ A-${\mathcal{L}_{d_{i,j}}}$&  65.7  &  41.6  &   40.3     \\\hline 
\end{tabular}
}
\caption{Success rate of different training methods using Deeplab backbone and 10 hours of demonstration on the regular traffic CARLA simulator.}\label{tab:6}
\end{table}

Besides,the Wasserstein loss is stabilized after 3$\times10^5$ steps. Training with more steps does not affect the performance until 5$\times10^5$ steps. Actually, based on our experiments for 10$\times10^5$steps, the curve can be stable. The window for training step of Wasserstein loss does not require careful tuning in our tested datasets.  

CARLA characterizes the approaches by average distance traveled between infractions of the following five types: opposite lane, Sidewalk, collision with static object, collision with car, collision with person.

CARLA offers a fine-grained evaluation of driving polices which characterize the approaches by the average distance between different collisions and more than 30\% off-line or off-road. The results are reported in Table \ref{tab:4}. Rather than test on the same town environment, we also test at a new town or new weather condition following the standard evaluation of CARLA. As expected, our method can largely improve these metrics and lead to a more safe driving system. By emphasizing the severity of misclassification of person, the average distance between two collisions with a person almost doubled in all of the testing cases. Besides, in \ref{tab:6}, we evaluate the success rate \cite{codevilla2019exploring} of different training methods using Deeplab backbone and 10 hours of demonstration on the regular traffic CARLA simulator. We can see that the Wasserstein loss can consistently improve the success rate.

Other than using our reinforcement learning framework to make the driving decision, we also evaluate our segmented results using an independent autonomous driving system. \cite{fennessy2019autonomous} propose to process the front view image in CARLA with Deeplab \cite{chen2017deeplab} to get a segmentation and then combine it with the depth camera and vehicle stats as state. We replace its vanilla Deeplab module with a fine-tuned one using Wasserstein loss or IAL. Following the experiment setting and evaluation metrics, we give the comparison in Table \ref{tab:5}. We use the prefix A to denote the adaptive ground metric learning. The improvements over Deeplab and IAL trained Deeplab indicate that our segmenter can offer more reliable and safe segmentation results for the driving system.

\section{Conclusions}

In this paper, we proposed a concise loss function for semantic segmentation in context of safe driving, based on the Wasserstein distance. The ground metric of Wasserstein distance represents the pair-wise severity and can be either predefined or learned by alternative optimization. The importance-aware problem can be a special case of our framework. Configuring a convex function of $d_{i,j}$ can further improve its performance. It has a simple exact fast solution in one-hot case and the fast approximate solution can be used for the conservative label in self-learning based unsupervised domain adaptation. We not only achieve the promising results in importance-aware tasks, but also improve the autonomous driving metrics in CARLA simulator significantly. For the feature work, we are planing to apply it to more advanced backbones and use real world evaluations to adjust the ground matrix \cite{zhou2019automated}.

\section{Acknowledgements}

This work was supported by the Jangsu Youth Programme [grant number SBK2020041180], National Natural Science Foundation of China, Younth Programme [grant number 61705221], and Hong Kong Government General Research Fund GRF (Ref. No.152202/14E) are greatly appreciated.

\bibliographystyle{ieee}
\bibliography{egbib2}

\end{document}